\documentclass{article}
\usepackage{collas2024_conference,times}
\usepackage{easyReview}

\usepackage{hyperref}
\hypersetup{
    colorlinks=true,
    linkcolor=red,
    filecolor=magenta,
    urlcolor=blue,
    citecolor=purple,
    pdftitle={Statistical Context Detection for Deep Lifelong Reinforcement Learning},
    }

 \usepackage{todonotes}
 \setuptodonotes{disable}

 \usepackage{array}
 
 \usepackage{makecell}

 \usepackage{amsfonts,amsmath,amssymb}
 \usepackage{algorithm,algorithmic}

 \usepackage{pgf, tikz, tikzscale, pgfplots}
 \usepgfplotslibrary{units}
 \newcommand{\mathdefault}[1][]{}

 \title{Statistical Context Detection for Deep Lifelong Reinforcement Learning}

 \author{Jeffery Dick\textsuperscript{1}, Saptarshi Nath\textsuperscript{1}, Christos Peridis\textsuperscript{1}, Eseoghene Benjamin\textsuperscript{1,2}, Soheil Kolouri\textsuperscript{3}, Andrea Soltoggio\textsuperscript{1}
 \\
 \textsuperscript{1} Loughborough University, United Kingdom\\
 \textsuperscript{2} The Alan Turing Institute, United Kingdom\\
 \textsuperscript{3} Vanderbilt University, Tennessee\\
 \texttt{j.dick@4open.science, \{s.nath, c.peridis, a.soltoggio\}@lboro.ac.uk}
}

 \collasfinalcopy
 
\begin{document}

\maketitle

\begin{abstract}
Context detection involves labeling segments of an online stream of data as belonging to different tasks. Task labels are used in lifelong learning algorithms to perform consolidation or other procedures that prevent catastrophic forgetting. Inferring task labels from online experiences remains a challenging problem. Most approaches assume finite and low-dimension observation spaces or a preliminary training phase during which task labels are learned. Moreover, changes in the transition or reward functions can be detected only in combination with a policy, and therefore are more difficult to detect than changes in the input distribution. This paper presents an approach to learning both policies and labels in an online deep reinforcement learning setting. The key idea is to use distance metrics, obtained via optimal transport methods, i.e., Wasserstein distance, on suitable latent action-reward spaces to measure distances between sets of data points from past and current streams. Such distances can then be used for statistical tests based on an adapted Kolmogorov-Smirnov calculation to assign labels to sequences of experiences. A rollback procedure is introduced to learn multiple policies by ensuring that only the appropriate data is used to train the corresponding policy. The combination of task detection and policy deployment allows for the optimization of lifelong reinforcement learning agents without an oracle that provides task labels. The approach is tested using two benchmarks and the results show promising performance when compared with related context detection algorithms. The results suggest that optimal transport statistical methods provide an explainable and justifiable procedure for online context detection and reward optimization in lifelong reinforcement learning. 
\end{abstract}

\section{Introduction}
Deep reinforcement learning algorithms \citep{sutton2018reinforcement,Mnih2015} for learning single tasks have become effective in many domains \citep{arulkumaran2017deep}. The extension to learning multiple sequential tasks is a focus of recent research in lifelong reinforcement learning (LRL) \citep{khetarpal2022towards}. Such effort is motivated by the observation that real-life scenarios could entail many sequential tasks. LRL borrows from the related field of lifelong learning (LL) \citep{thrun1998lifelong,silver2013lifelong,kudithipudi2022biological,soltoggio2024collective} that more generally research machine learning (ML) algorithms that perform well with different ML paradigms when data distributions change over time. To aid the LL algorithm, ML research to detect, categorize, or label different data distributions is becoming increasingly more relevant \citep{kolouri2019sliced,kim2022continual}

Different approaches to LL \citep{parisi2019continual,deLange2022} have proven effective in a variety of different domains. Such approaches can be grouped into weight plasticity and regularization methods, replay methods, and structural adaptation methods. Weight plasticity and regularization approaches include EWC~\citep{kirkpatrick2017overcoming}, SCP~\citep{kolouri2019sliced}, SI~\citep{zenke2017continual}, MAS~\citep{aljundi2018memory}, and ANCL~\citep{kim2023achieving}. These methods work by anchoring learning around the important parameters of previous tasks. Therefore, these approaches require task labels, or at least the task change time, to consolidate the parameters when required.  One exception is the task-free continual learning protocol by \citet{aljundi2019task} that performs consolidation when the agent's performance is stable, effectively implementing an implicit detection of good performance on a task (Section~\ref{sec:rel-other}).

Examples of memory and replay approaches include REMIND~\citep{hayes2020remind}, DGR~\citep{shin2017continual}, and LwF~\citep{li2017learning}. While most of these methods focus on classification, experience replay has been used in the reinforcement learning setting~\citep{mnih2015human} with a particular focus on continual learning in CLEAR~\citep{rolnick2019experience}. Under certain conditions, these methods may not require task labels, but in such a case, replay buffers require unbounded size to store experiences from all tasks. 

Methods that adapt the structure of the network to learn multiple tasks include multi-head approaches~\citep{YosinskiCBL14}, Progressive NNs~\citep{rusu2016progressive}, XdG~\citep{masse2018alleviating}, PathNet~\citep{fernando2017pathnet}, mask based approaches~\citep{ben2022lifelong, wortsman2020supermasks}, and HAT~\citep{serra2018overcoming}. In these approaches, weights are either added to the network to scale to new tasks or different weights are activated for dealing with different tasks. Task labels are particularly useful with structural adaptation methods to associate particular structures, e.g., network masks, with tasks.

Task changes in RL can be categorized as changes in the input distribution, changes in the transition function, or changes in the reward function. Methods to detect changes in input distributions have been developed in the field of novelty and out-of-distribution detection with applications in LL \citep{geisa2021towards,van2022three,aljundi2022continual,liu2022wasserstein}. One interesting property of LRL is that changes in the reward function cannot be detected only by looking at the distribution of the observations under a random policy. In such cases, the policy can determine or change the distribution, effectively making it more difficult to determine the current task. In this paper, we address specifically this case.

The algorithm proposed in this paper aims to detect changes in deep RL tasks and match the best policy for each task. It is designed to operate online by measuring distances between experiences in the data stream, represented by suitable latent spaces in the network architecture. Using statistical methods, the proposed algorithm determines whether the task has changed. To do so, mechanisms are introduced to achieve a desired low level of false positives, to account for distribution changes under evolving policies, and to re-detect changes seen in the past. Task similarities are measured via the Sliced Wasserstein Distance (SWD) which is subsequently used with the Kolmogorov-Smirnov (KS) statistical test to determine task changes. The resulting method is named \emph{Sliced Wasserstein Online
 Kolmogorov-Smirnov (SWOKS)}.
Simulations are conducted on the CT-graph \citep{soltoggio2023configurable},  Minigrid \citep{chevalier2021minimalistic} and Half-Cheetah \citep{todorov2012mujoco} benchmarks, and compared with the baselines Task-Free Continual Learning (TFCL) \citep{aljundi2019task}, Model-Based RL Context Detection (MBCD) \citep{alegre2021minimum} and Replay-based Recurrent Reinforcement Learning (3RL) \citep{caccia2023task}.

\section{Related Work}
The effort to detect novel distributions and changing tasks has seen developments in a number of research areas that are reviewed in this section. The following approaches provide the foundations and motivate the novel method proposed later in section \ref{sec:kolsmi}.

\subsection{Novelty and Out Of Distribution Detection}
\label{sec:ood}
In the absence of labeled data, out-of-distribution (OOD) detection methods such as \citet{lee2018simple,haroush2021statistical,de2016minas} have been developed to detect novel distributions in classification. These methods classify individual samples as in or out of distribution. Similarly, novelty detection ~\citep{sedlmeier2019uncertainty} aims to determine when data no longer fits an expected distribution. Such approaches may prove useful in LL ~\citep{aljundi2022continual} by providing confidence values on distributions of incoming data. One example is the inductive conformal prediction (ICP) framework~\citep{papadopoulos2008inductive} that provides a statistical estimation, or confidence, of the correctness of a predictive function. ICP learns such estimates via training, calibration, and validation phases, which limits its applicability to LL.  The max-Simes-Fisher (MaSF) algorithm \citep{haroush2021statistical} frames OOD detection in DNNs as a statistical hypothesis testing problem. It uses evidence from the entire network to generate $p$-values for each sample. 

\subsection{Context Detection} 
Context Detection methods are designed to detect changes in context or task in reinforcement learning (RL). Early examples include approaches such as~\citet{da2006dealing,da2006improving} that can learn and label different tasks for tabular RL algorithms by tracking prediction error rates of partial models. 
For more challenging approximate methods, \citet{kessler2022same} introduced OWL that learns task labels during a preliminary training phase.

\subsubsection{Model Based RL Context Detection}
In the field of model-based RL, \citet{alegre2021minimum} introduced an algorithm (MBCD) for online context detection with approximate methods that do not require pre-training.  By utilizing an ensemble model of neural networks, MBCD estimates a model of the environment, including the probability of given state-action-reward-state (SARS) tuples appearing. The SARS tuples are associated with an estimated probability, emulating the tabular methods applied in discrete state (PO)MDP environments. MCUSUM is used to determine which task is most likely: the current task, a previous, a seen task, or a brand new task. One objective of MBCD is task detection with few samples. This objective is achieved by training simultaneously a Bayesian Neural Network (BNN) and a policy network. The BNN is trained on a rolling window of the agent's history, and at each time step predicts what the environment's output should be. One disadvantage of the method is that the training of the BNN can be computationally expensive. 

\subsection{Other approaches to task detection in lifelong learning}
\label{sec:rel-other}

In an early approach to LRL, \citet{brunskill2013sample} introduced a multi-task learning algorithm for use in tabular methods (i.e., prior to the introduction of deep approximate RL). The algorithm, however, requires a pre-training phase to learn the parameters of different MDPs and applies clustering to determine which knowledge is beneficial in new tasks.

Task-free LL approaches such as \citet{aljundi2019task, ye2022task} learn from a non-stationary data stream without an explicit task boundary given. The approach in \citet{aljundi2019task} consolidates knowledge when the agent's performance is stable, thus avoiding the problem of task detection. One drawback to many task-free LL approaches is the reliance on a single policy network to learn multiple tasks, which is unable to deal with interfering tasks if labels are not provided to the input \citep{kessler2022same}.

Promising novel approaches are based on metrics that measure distances between datasets. In \citet{liu2022wasserstein}, Wasserstein embeddings (WE) are used that map a set of high-dimensional data points to a lower-dimensional Euclidean space. The distance between embeddings of different sets approximates the Wasserstein distance (WD) between the sets and can be used to measure the similarity between sets. These types of approaches are an essential step to perform statistical tests and assess whether an incoming stream of data belongs to the same task or to a different task.

\section{Preliminaries}
\subsection{Task and Environment}
In RL, at each time step $t \in \mathbb{N}$, an agent is exposed to an observation $o_t \in {\Gamma}$ and reward $r_t \in \mathbb{R}$. The observation is a function of the state,  $o_t = {O}(s_t) \in {\Gamma}$. The agent can influence the next state encountered ($s_{t+1}\in S$) by taking actions at each time step, $a_t \in {A}$, with probability given by the transition function, ${T}:({S} \times {A}) \rightarrow ({S})$. The transition is associated with a reward given by the reward function, $r_t={R}(s_t)$. In LRL environments, tasks are added to the environment definition. Each task $k \in K$ may be associated with its own
$S_k,A_k,T_k,R_k,\Gamma_k$ and $O_k$.

\subsection{Sliced Wasserstein distance}
The Wasserstein distance (WD) is a measure of the distance between two probability distributions over a given metric space. It originates from the field of optimal transport, where the goal is to find the most efficient way to transform one distribution into another. It has gained increased popularity recently for measuring distances in datasets, for use in RL, classification learning, and generative image models~\citep{arjovsky2017wasserstein}.

There are many notions of dissimilarity for probability measures, among which are $\phi$-divergences, integral probability metrics, and optimal transport-based distances (e.g., Wasserstein distances). In our application, we have specific desiderata for such a dissimilarity measure: (i) it should be able to measure dissimilarities between probability measures with disjoint supports; (ii) it should have a small sample complexity; (iii) it should be computationally efficient. The sliced-Wasserstein distance is one such measure that satisfies all these conditions, whereas alternative distances do not. For instance, Wasserstein distances exhibit poor sample complexity and are not computationally efficient, while $\phi$-divergences are unsuitable for comparing distributions with non-overlapping supports.

Given two datasets $D_{1}, D_{2}$ of equal size, assume each data point has equal significance. Then, the WD between those two datasets is given by:
\begin{equation} \label{eq:Wass}
      W(D_1,D_2) = \inf_\sigma \sqrt{\sum_{i=1}^{n}\|D_{1,i}-D_{2,\sigma(i)}\|^2}\quad,
\end{equation}
where {$\sigma:[1,n]\cap\mathbb{N}\rightarrow[1,n]\cap\mathbb{N}$} is a permutation~\footnote{The WD can use a non-bijective mapping: we use one in this case due to both datasets containing the same number of equally weighted points.}. Solving Eq.\ \ref{eq:Wass} can be computationally expensive for large data sets. For this reason, the sliced Wasserstein distance (SWD) \citep{rabin2012wasserstein, bonneel2015sliced, kolouri2019generalized} is often used to reduce the computational requirements whilst still providing a useful approximation to the true WD. The SWD provides an estimation of the true WD via radon transforms:
\begin{equation}
    SW(D_1,D_2) = \int_{\mathcal{S}}W(D_1^\theta,D_2^\theta)\; d\theta\quad,
\end{equation}
where $\mathcal{S}$ is the unit sphere, and $D^\theta$ is the radon transform of $D$ at the angle of $\theta$. Due to the one-dimensional nature of $D_i^\theta$, and given that the optimal transport map is cyclically monotonous (i.e., in one dimension, it is monotonically increasing), the optimal permutation $\pi$ can be obtained by sorting the sliced samples, $\{D^\theta_{k,i}\}_{i=1}^{N}$. By choosing a high number of sample projections, a high accuracy can be attained.

\subsection{Kolmogorov-Smirnov statistical test}
The Kolmogorov-Smirnov (KS) statistical test is designed for use on non-parametric distributions. It compares two empirical cumulative distribution functions (ECDFs) and determines whether these samples are likely to be drawn from different distributions (alternative hypothesis). The null hypothesis is that the samples are drawn from the same distribution. We specifically use the one-sided KS test.
For two ECDFs $X_1$ and $X_2$, the one-sided KS statistic is defined as
  \begin{equation} \label{eq:KolSmi}
      KS(X_1,X_2) = \sup_{x}
      \left(
      \text{P}[X_1<x]
      -
      \text{P}[X_2<x]
      \right)
      \quad.
  \end{equation}
The null hypothesis that the sample distributions are drawn from the same underlying function is rejected at confidence level $\alpha$ if
  \begin{equation}
  KS(X_1,X_2) >
  \sqrt{\frac{-0.5\left(\left|X_1\right|+\left|X_2\right|\right)
  \times\text{ln}(0.5\alpha)}{\left|X_1\right|
      \times\left|X_2\right|}}
  \quad.
  \end{equation}

Enhancements to the K-S have been developed to overcome limitations such as the applicability to continuous distributions only, e.g., the Anderson-Darling \citep{10.1214/aoms/1177729437} or the Cramer Von-Mises \citep{anderson1962distribution} tests, which could be considered a refinement of the proposed method.

\subsection{Network Masks}
Network masks are structural adaptation methods for lifelong learning. Their use in supervised learning is demonstrated in \citet{wortsman2020supermasks,kim2022continual}. Masks in reinforcement learning have also shown promise due to their ability to combine knowledge from multiple tasks \citep{ben2022lifelong,nath2023L2D2-C}.
Masking approaches work by combining a network with constant parameters $\theta$ with a mask that identifies a subset of $\theta$. Multiple such masks $\{m_1, m_2, \ldots, m_n\}$ with  parameters $\phi_i$ can be used to learn multiple policies. Due to the separation of masks, interfering tasks can be learned by different masks.
When a particular task $i$ is identified as the current task, the scores for mask $m_i$ are applied and trained using edge-popup~\citep{bengio2013estimating}. Details of the implementation of the masking approach follow those in \citet{ben2022lifelong} and are further expanded in the Appendix.

\section{Sliced Wasserstein Online
Kolmogorov-Smirnov (SWOKS)}
Let $K = \{{k}_1, {k}_2, \ldots, {k}_n\}$ be the set of tasks available in the environment in which each task has a corresponding ${T}_k$, ${R}_k$, and ${O}_k$ function. 
Let $k_t$ be the task governing the environment at time step $t$, and $z' \in Z$ be the label predicted by an agent at that time. 
The algorithm attempts to minimize the sum of differences between predicted labels and true tasks (where $\delta$ is the \textit{Dirac delta} function):
   \begin{equation}\label{eq:obj}
    {E} = \min_F \left( \sum_{t \in T} \delta \left( F(z')-k_t \right) \right)\quad.
   \end{equation}

The correct identification of the present context allows a LL system to train multiple policies $\Pi = \{\pi_1, \pi_2,...,\pi_n\}$ with different subsets of the data stream with the aim of maximizing the {infinite horizon return at each timestep}. 

Maximizing the expected return relies upon both correct task labels and the successful training of each policy.
The key aspects of the algorithm are (1) the tuning of the false positive and false negative rates for the specific problem domain (Section~\ref{sec:kolsmi}); (2) the ability to re-detect a previously seen task (Section~\ref{sec:taskchange}); (3) the ability to match a subset of the data stream to the correct policy for each task (Section~\ref{sec:taskchange}). 

\begin{figure}
\centering
 \includegraphics[width=0.9\textwidth]{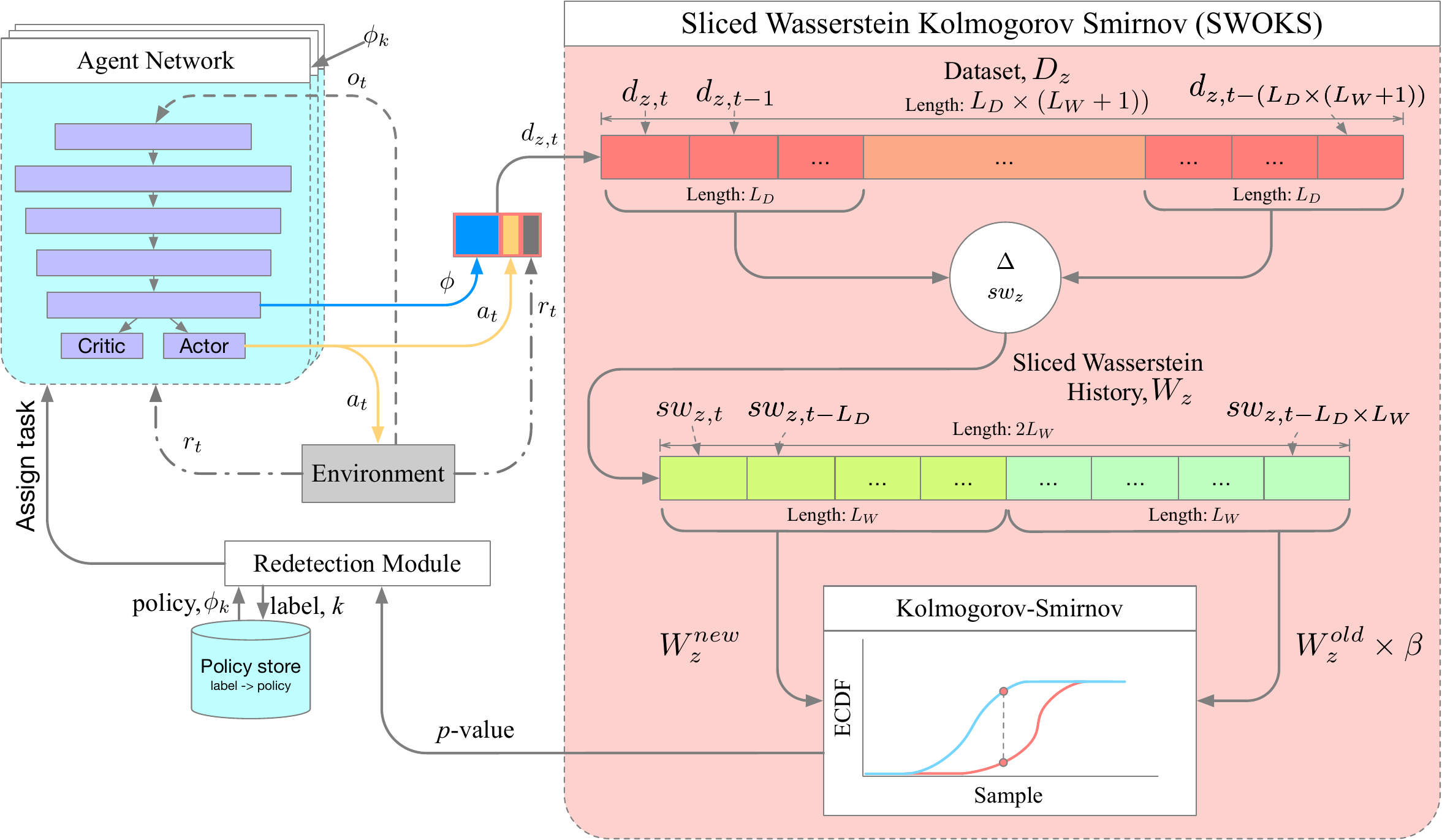}
   \caption{Graphical representation of the SWOKS architecture. }
\label{fig:datapoint}
\end{figure}

SWOKS uses the most recent $L_D$ data points in the stream as reference set $(D_0)$ to compute a distance with an older set $D_z^{old}$, $L_D \times L_W$ time steps in the past. $L_W$ is the number of sets, corresponding to distance measurements, that can be taken over a sliding window buffer of length $L_D \times (L_W + 1)$. Thus, $D_{z'} \in \mathbb{R}^{\left(L_D\times (L_W +1)\right),\left( |\phi|+2\right)}$ where $d_{0,t} \in \mathbb{R}^{\left(|\phi|+2\right)}$ is a single data point in $D_0$, and contains the concatenation of $\sqrt{len(\phi|t)}\times r_t, a_t,$ and $\phi_t$ (Fig.\ \ref{fig:datapoint}). In summary, 
\begin{equation}\label{eq:TaskWass}
    sw_{z,t} =  SW( D_{0}, D_{z}^{old})\quad,
\end{equation}
is computed for each new set $D_0$, thus at a frequency of $L_D$ time steps.

\subsection{Kolmogorov-Smirnov calculation and adjustments}\label{sec:kolsmi}
The Kolmogorov-Smirnov test is performed on the set of data points derived from Eq.\ \ref{eq:TaskWass} at every $L_W$ time steps, comparing the most recent distances $sw$ from indices $[t, \ldots, t-L_W]$ and an older set of distances $sw$ at indices $[t-L_W-1, t-(2\times L_W)]$. However, the application of the standard test does not lead to useful results due to the multiple testing problem \citep{weisstein2004bonferroni,benjamini1995controlling} that increases the probability of making at least one Type I error when performing numerous tests. This corresponds in our case to a false detection of task change. Various methods for adjustments to reduce the rate of Type I errors have been proposed in literature \citep{benjamini1995controlling}.
Bonferroni correction alone is not useful for SWOKS as the $\alpha$ adjustment must scale to the number of tests performed, which in our case is unbounded.
Preliminary tests (not shown) suggested a correction procedure that works well in SWOKS: a multiplicative increase ($\beta$) of the reference SWD ($W_z^{old}$) to reduce the frequency at which small variances in SWD hold statistical significance. The standard setup has a significance level $\alpha=0.001$ and a $\beta = 1.1$ upward adjustment for the reference SWD $W_z^{old}$. Future experimental tests could establish the exact relationship between such settings and the precise rate of Type I and Type II errors. However, in the set of experiments we ran across all proposed benchmarks, we observed that such adjustments are robust enough to reliably detect task changes. One strength of the algorithm is that, if experiments on one particular benchmark fail due to excessive Type I or II errors, the $\beta$ parameter can be adjusted accordingly.

\subsection{Task change detection and re-detection challenges}\label{sec:taskchange}

When the $p$-value generated by the KS calculation drops below the significance threshold parameter $\alpha$, the most recent data is assumed to be produced by a new, different distribution. If this is the first task change detected by SWOKS, a new task label is created to track the new detected task. Otherwise, SWOKS must decide whether a previously seen task label fits the current data, or whether a new task label should be created (Algorithm 2). 

Task re-detection poses the following challenges. 
\begin{enumerate}\setlength{\itemsep}{0pt}
\item The policy under which a task was experienced in the past is an integral part of the data distribution. 
\item A consequence of the previous point, a policy that undergoes learning will also lead to a shifting distribution of the observations, the latent spaces, and the rewards. 
\item The KS test reveals when two distributions are different for detecting new tasks, but {no statistical method offers a} test to rematch previously seen distributions.
\item The exact data point at which a distribution change might not be known because each test performed corresponds to a set of observations rather than a single data point. 
\end{enumerate}

To address challenge 1 when attempting to re-detect a given task, the agent must deploy the policy that was used when originally experiencing that task. The solution to the swapping policy adopted here is based on recent advances in the use of modulating masks in LRL~\citep{ben2022lifelong}. Masks are compact representations of policies that can be applied to network backbones to learn multiple tasks. SWOKS reproduces the LRL implementation proposed in~\citet{ben2022lifelong}. 

To address challenge 2, two solutions are implemented. The first is to enable task detection only after an initial learning phase when engaging with a new task. We set a value \emph{stablePhase} $= 50,000$ time steps. A second solution is to use RL algorithms that perform gradual policy updates, e.g., TRPO~\citep{schulman2015trust} or PPO~\citep{schulman2017proximal}.

To address challenge 3, we consider the KS test results after each policy assessment in the loop through policies devised to address challenge 1. If the $p$-value is greater than the significance threshold, it means the data is not different enough to assume it came from a different task. Whilst this test does not guarantee the tasks are the same, if the tasks appear indistinguishable it is reasonable to treat them as the same task.

To address challenge 4, SWOKS implements a roll-back mechanism. The current policy, encoded as a network mask, and trained via PPO, is backed up every 50 iterations. Upon task detection, the current mask is considered corrupted with data from the new task, and the previously backed-up network is stored in association with the previous task. In addition, a SAC implementation is used to benchmark SWOKS in continuous environments. In this case, we used the same setup as in MBCD \cite{alegre2021minimum}. All the steps outlined in this section are summarized in Algorithms 1 and 2.
\begin{figure}[t]
    \centering
\begin{minipage}[t]{0.5\textwidth}
     \begin{algorithm}[H]
       \label{alg:WOKS}
       \caption{SWOKS main loop}
       \begin{algorithmic}
         \STATE $z'\leftarrow$1 {\quad //} current task
         \STATE $t\leftarrow$1 {\quad //} time step
         \STATE $Z\leftarrow$[1]
         \STATE $last\_z\_change \leftarrow 1$
         \WHILE{true}
         \STATE $t+=1$
         \STATE $D_{z'}.\mathrm{FIFOupdate}([\phi_t, \;a_t, \;r_t])$
         \IF{$L_D \% t==0$}
         \STATE $W_{z'}.\mathrm{FIFOupdate}(sw(D_{z'}^{new})$ 
         \STATE $K_{z'}\leftarrow KS(W_{z'}^{new}, W_{z'}^{old}\times \beta)$
        \IF{$K_{z'}<\alpha$}
         \STATE {Re-detection Algorithm}
         \ENDIF

         \ENDIF
         \ENDWHILE
       \end{algorithmic}
     \end{algorithm}
   \end{minipage}%
   \begin{minipage}[t]{0.5\textwidth}
     \begin{algorithm}[H]
       \label{alg:Testing}
       \caption{Re-detection}
       \begin{algorithmic}
         \IF{$t - last\_z\_change < stablePhase$}
         \STATE Abort testing procedure
         \ENDIF
         \FOR{$z\in Z \setminus z'$}
         \STATE $D^*_0$ = test(policy($z$))
         \IF{$p\mathrm{-val}_{task}>\alpha$}
         \STATE $D_0\leftarrow D^*_0$
         \STATE $z' \leftarrow z$
         \STATE Exit()
         \ENDIF
         \ENDFOR
         \STATE $Z$.append($z_{|Z|+1}$) 
         \STATE $z'\leftarrow z_{|Z|+1}$
         \STATE $last\_z\_change \leftarrow t$
       \end{algorithmic}
     \end{algorithm}
   \end{minipage}
   \end{figure}
The source code for reproducing the simulations is available at \url{https://github.com/JupiLogy/swoks}. 

\section{Experiments}
Simulations are conducted on three benchmarks: CT-graph~\citep{soltoggio2023configurable}, Minigrid~\citep{chevalier2021minimalistic}, and Mujoco Half-Cheetah. The CT-graph and Half-Cheetah environments have task-independent observations, i.e., different tasks have the same observation space. The Minigrid environment has observations that may vary slightly across tasks. SWOKS is implemented in combination with modulating masks and PPO for the environments CT-graph and Minigrid, and with SAC for the continuous environment Half-Cheetah.

The Mujoco Half-Cheetah environment is set up with four different tasks, performed in sequence, as in \citet{alegre2021minimum}. Observations given by the environment are a vector of seventeen real numbers, giving information about the location, rotation, and velocity of different sections of the simulated robot. The first task is the default ``normal'' Half-Cheetah environment. The other three tasks are ``joint-malfunction'', ``wind'', and ``velocity'' tasks respectively. These task changes affect the physics of the environment, i.e., the transition function, but not the rewarding conditions or observations made by the agent. SWOKS, MBCD, and 3RL are evaluated in the Half-Cheetah environment. Further details are provided in the Appendix.

The Configurable Tree graph (CT-graph) environment is a configurable tree graph designed for testing LRL algorithms due to its feature of automatically creating multiple tasks with measurable levels of complexity, sparsity of reward, and similarities among tasks. In the setup for this study, we used four tasks with the same observations and transition function, but different reward functions. This environment is used to evaluate SWOKS and the TFCL algorithm \citep{aljundi2019task}.

The Minigrid environment is a lightweight grid-world environment designed to test the generalizability and adaptability of LRL algorithms due to its scalability to generate various permutations of tasks, enabling varying complexities and similarities. The environment employs sparse rewards. In this study, we used three seed variations of the SimpleCrossingS9N2-v0 environment, maintaining certain similarities between tasks. This environment is used to evaluate SWOKS and the TFCL algorithm.
 
\subsection{SWOKS on interfering tasks and observation space changes}

{To test the resilience of SWOKS to task interference, we compare its performance to TFCL. Both algorithms are equipped with the same network structure, though TFCL does not make use of the modulating masks due to lacking a task change detection mechanism. TFCL uses a ``prioritized keeping" replay buffer in order to retain knowledge on old tasks. When compared with SWOKS on the 4-task CT-graph environment, we observe that the TFCL is unable to maintain performance on all tasks despite the continual learning mechanism. This inability is caused by the same observation and transition functions, and different-reward properties of the four tasks in this benchmark creating interference between tasks. SWOKS, on the contrary, appears to learn all tasks. It is important to note that the modulating masks method requires task labels or in this case the automated SWOKS mechanism, to enable the algorithm to select the correct mask for each task.}

\begin{figure}
 \centering
 \begin{tabular}{cc}
   a) & \includegraphics[width=0.9\textwidth]{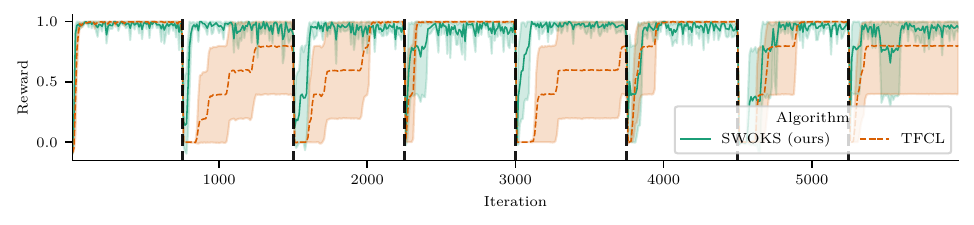}\\
    b) & \includegraphics[width=0.9\textwidth]{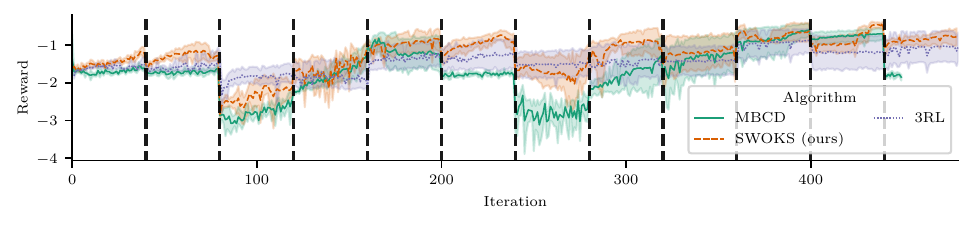}

 \end{tabular}
     \caption{Reward achieved over sequences of tasks.
     a) Average reward over 5 seeds for SWOKS and TFCL in the CT-graph environment. The sequence of tasks from 1 to 4 is seen twice in the order 1-2-3-4-1-2-3-4. The rolling average over 10 iterations is plotted for easier viewing.
     b) SWOKS, MBCD and 3RL are tested on the Half-Cheetah environment. Task changes occur every 40000 timesteps (40 iterations).}
   \label{fig:tfcl}
 \end{figure}

\begin{figure}
    \centering
    \includegraphics{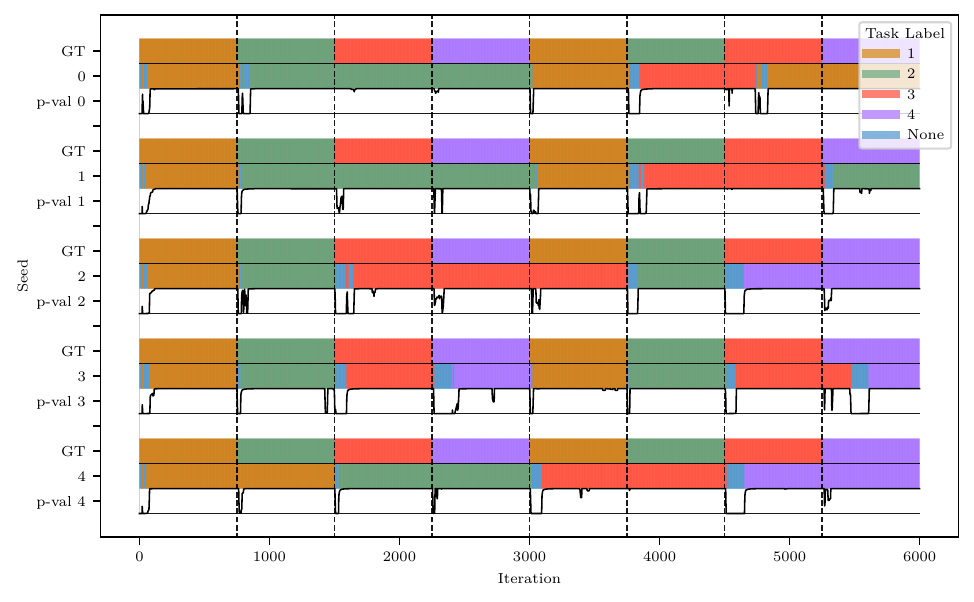}
    \caption{Detected task over time for 5 seeds of the SWOKS agent performing in the CT-graph environment. The ground truth (GT) task is compared with the task label generated by each agent. Task labels given in this graph are from the same run as Figure~\ref{fig:tfcl}. The label "None" corresponds to the agent having a low $p$-value for all previously seen tasks.}
    \label{fig:swoks-detected}
\end{figure}

The reward achieved by SWOKS and TFCL in the CT-graph is shown in Figure~\ref{fig:tfcl}(a). The $p$-values and detected task labels are found in Figure~\ref{fig:swoks-detected}.

\begin{figure}[t]
\centering
\begin{tabular}{cc}
(a)&\includegraphics[width=0.8\textwidth]{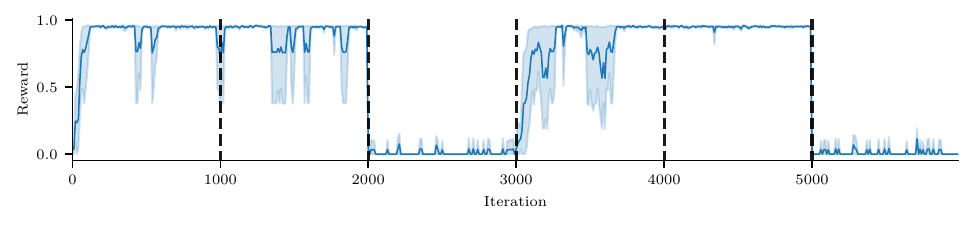}\\\vspace{-10pt}
&\includegraphics[width=0.9\textwidth]{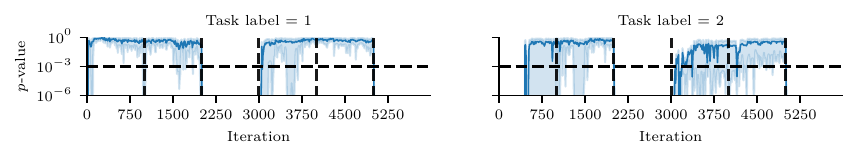}\\
(b)&\includegraphics[width=0.45\textwidth]{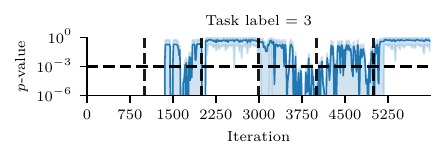}\\\vspace{-10pt}
\end{tabular}
\caption{Analysis of learning dynamics on the Minigrid environment for two seed runs. Task changes occur every 1000 iterations, between 3 tasks (details in appendix). The three tasks are each seen twice in the order 1-2-3-1-2-3. (a) Average reward: the first two tasks are learned, but the third one fails. (b) The $p$-values are high (above $0.001 = 10^{-3}$) when a task is detected. We observe that task 2 is mistakenly identified as task 1 (top left graph) but without affecting the performance. Task 3 is also correctly identified despite the agent failing to learn it.}
\label{fig:minigrid}
\end{figure}

SWOKS is designed to detect tasks in difficult cases when changes occur in the reward or transition functions. However, changes in the observation space may also be detected. In the following simulations, SWOKS is tested on three tasks in the Minigrid environment specifically to assess the ability of SWOKS to deal with variations in the input distribution. Fig.\ \ref{fig:minigrid}(a) shows that tasks 1 and 2 are learned but task 3 fails to learn. Nevertheless, SWOKS identifies the third task as an additional one as shown in Fig.\ \ref{fig:minigrid}(b).

\subsection{SWOKS on continuous action spaces}

SWOKS is compared to \citet{alegre2021minimum}'s MBCD algorithm and \citet{caccia2023task}'s 3RL algorithm. This comparison is made in the Half-Cheetah environment with the same task setup as used in~\citet{alegre2021minimum}. MBCD is used without simulated replay to work under the same assumptions on input data as SWOKS and 3RL. 
The average reward over 5 seeds for each algorithm is plotted in Figure~\ref{fig:tfcl}(b). While 3RL reaches a plateau with a policy that works fairly well for all tasks, MBCD and SWOKS each apply new network masks when new tasks are detected, allowing for more robustness against interfering tasks. MBCD is known to assign the same task label for different tasks that may be similar. This allows for knowledge transfer between similar tasks, but in these experiments resulted in lower performance.

\subsection{Analysis of $\beta$ and $p$-values}\label{sec:params}

 \begin{figure}
     \centering
     \includegraphics[width=0.9\textwidth]{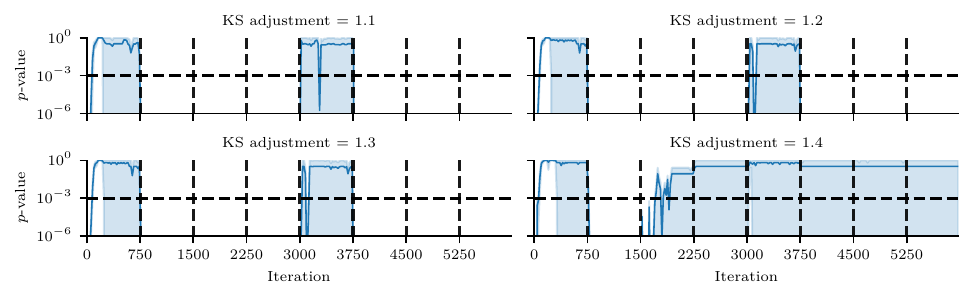}
     \caption{Plot over time of $p$-values for task label 1 and for different values of $\beta$ in the CT-graph benchmark. The curriculum is formed of tasks 1-2-3-4-1-2-3-4. We can observe that task 1 is correctly identified for all settings, but with $\beta=1.4$, SWOKS also believes tasks 2, 3, and 4 to be tasks 1. A rolling mean is taken over a sliding window of size 20 for readability.}
     \label{fig:beta-anal}
 \end{figure}

The KS $p$-values generated by SWOKS are essential to its task detection capabilities. The plots in Figure~\ref{fig:minigrid} show that the $p$-values are high for the matching task, and lower for the other tasks, when the task changes are correctly identified.
The KS adjustment, $\beta$, is a parameter introduced to reduce the number of false positives detected by SWOKS. An analysis of different values of $\beta$ on the task classification of SWOKS is seen in Figure~\ref{fig:beta-anal}.

\section{Discussion}

The proposed SWOKS algorithm introduces hyper-parameters that address fundamental assumptions on the dynamics of context detection and task change. Namely, how frequently tasks change over time, how many SAR data points are required to draw statistically significant conclusions, and what types of variations can be observed in the online data stream. Such aspects are highly domain-dependent, and it makes sense that tunable parameters should be used. SWOKS can be tuned with different $L_D$ and $L_W$ parameters and KS adjustment $\beta$.

Environments with many different states require larger $L_D$ and $L_W$ because different tasks can be differentiated only after sampling many observations over a prolonged period of time. $L_W$ determines the size of SWD samples, which are highly compressed representations of differences between sets of data. Thus, increasing $L_W$ may provide a computationally efficient way to increase the statistical power of the test provided that $L_D$ is sufficient to capture a minimum set of observations.

The KS adjustment $\beta$ parameter is an important feature of the algorithm to tune the type I error rate. As $\beta$ depends on the distribution of the data within SWOKS, using an adaptive $\beta$, dependent on the standard deviation of the data, could be investigated in future research in order to reduce the number of hyper-parameters of SWOKS.

Future work could extend SWOKS with a hyper-parameter analysis and its sensitivity to variation of the adjustments. Although such hyper-parameter tuning is a limitation of the approach, we note that, by means of such a tuning, SWOKS can be set to target a specific accuracy in the detection, providing a statistically motivated decision-making mechanism that can be set to address domain and problem-specific targets. 

\subsection{Scalability to many tasks}

One apparent limitation in same-observation-different-reward tasks is the requirement for policies to be tested in sequence to determine whether the new task is already known. It is important to note that this is not a limitation of SWOKS, but an intrinsic property of the problem. Assume a 4-task problem in which an agent in a square arena (e.g. Minigrid) needs to reach four different corners. After learning all four tasks, if the task changes again to one unknown corner, the only way to discover where the reward is located is to try all policies, i.e., explore all corners. However, if the distribution of observations helps in detecting tasks (i.e. via cues that lead to different distributions), SWOKS could be augmented with a clustering solution (e.g., like in \citet{brunskill2013sample} but still using the SWD (\citet{liu2022wasserstein} for high dimensional observations), to narrow the search to only specific policies. Such an approach could find the set of nearest policies prior to executing them.

\section{Conclusion} 
A novel algorithm, SWOKS, is introduced to detect tasks in deep RL where subtle changes involve transition functions or reward distributions. SWOKS exploits SWD to extract compressed representations of task changes and performs KS tests to provide task change detection and re-detection based on statistically meaningful decisions. The simulations suggest that the approach is a promising tool that can be used in combination with a policy gradient method (PPO) to optimize an overall measure of performance for an RL agent learning a sequence of tasks. Further tests are required to assess the flexibility of the method on a larger variety of benchmarks, including, e.g., continuous action spaces. While this study provides only a proof-of-concept, it also encourages further studies on the key idea of targeting desired error rates in task detection by means of statistical tools.

\bibliographystyle{icml2024}
\bibliography{combinedbib.bib}

\appendix

\section{Implementation details}
 \subsection{SWOKS setup and hyper-parameters}
Table \ref{tab.hyperparameters} provides a breakdown of the specific SWOKS hyper-parameters used in each of the experiments provided in this paper. We provide a brief explanation of each of these hyper-parameters and their effects on the operation of SWOKS.

    \begin{itemize}
        \item \textbf{History length, $L_{D}$:} Determines the length of the data taken from the beginning and end of the Dataset, $D_{z}$. A higher value provides more statistical power (i.e., the p-value is more likely to reach 0) and better stability, at the cost of slower detection speed.
        \item \textbf{Wasserstein history length, $L_{W}$:} Similar to the history length, this determines the length of the data taken from the beginning and end of the Sliced Wasserstein history, $W_{z}$. A higher value also increases the statistical power (with more effectiveness), at the cost of slower detection speed.
        \item \textbf{Significance threshold $\alpha$:} The value of $\alpha$ determines the base sensitivity of the p-value. A higher value triggers a task change at a higher p-value with a very minor effect. Very low values can cause SWOKS to be unable to detect task changes if the statistical power of the observations provided are not high enough to reach such small values.
        \item \textbf{KS adjustment $\beta$:} An adjustment value is used to stabilize the sensitivity of the task change detection. A higher value improves robustness to small oscillations in the $p$-value.
        \item \textbf{Stable phase duration:} A duration of n-steps to determine how long to wait before detecting task changes. This is necessary due to $p$-values dropping early on in the task-learning process as changing policies leads to exploration.
        \item \textbf{Model backup frequency:} Two consecutive backups of the model history are saved at specified intervals to ensure that if a task change detection is late, then the model of the previous task is not affected by the new task.
    \end{itemize}

    \begin{table}[b]
    \begin{center}
    \begin{tabular}{|p{5cm}|p{5cm}|}
    \hline
    Hyper-Parameter & Value\\
    \hline
        History length, $L_{D}$ & 240\\
        SWD history length, $L_{W}$ & 125\\
        Significance threshold, $\alpha$ & 0.001\\
        KS adjustment, $\beta$ & 1.1\\
        Stable phase duration & 50000 (100000 for MiniGrid)\\
        Model backup freq. (iterations) & 50\\
    \hline
    \end{tabular}
    \caption{Lists the SWOKS-specific hyper-parameters used in the experiments conducted in this paper.}\vspace{8pt}
    \label{tab.hyperparameters}
    \end{center}
    \end{table}

 \subsection{Network setup and hyper-parameters}
 SWOKS and TFCL are used with PPO for the CT-graph and Minigrid experiments. Table~\ref{tab.ppohyperparameters} shows the hyper-parameters used for the PPO network for each of these experiments. Table~\ref{tab.model} describes the network structure for these experiments in Minigrid, which is the same structure used in \citet{kessler2022same}.

 \subsubsection{Modulating masks}
 Continuous masks are used as in~\citet{ben2022lifelong}, and are not discretized at any point. Upon new task detection, for the CT-graph and Minigrid environments, the new masks are initialized randomly. For the Half-Cheetah environment, each newly created mask is created as a duplicate of the previously seen mask, allowing for better forward transfer of task knowledge. Combinations of modulating masks may provide more optimized mask initiation policies, however, this is not applied to the experiments in this paper.
 
    \begin{table}[H]
    \begin{center}
    \small
    \begin{tabular}{|p{5cm}|p{2.5cm}|p{2.5cm}|p{2.5cm}|}
    \hline
    Hyper-Parameter & CT-graph & Minigrid\\
    \hline
    Learning rate & 0.007 & 0.007 \\
    CL preservation & supermasks & supermasks \\
    Number of workers & 4 & 4 \\
    Optimizer function & RMSprop & RMSprop \\
    Discount rate & 0.99 & 0.99 \\
    Entropy weight & 0.01 & 0.01 \\
    Rollout length & 128 & 128 \\
    Optimization epochs & 5 & 5 \\
    Number of mini-batches & 64 & 64 \\
    PPO ratio clip & 0.1 & 0.1 \\
    Iteration log interval & 1 & 1 \\
    Gradient clip & 5 & 5 \\
    Max iterations (per task) & 750 & 1000 \\
    \hline
    \end{tabular}
    \caption{PPO agent hyper-parameters used for CT-graph and Minigrid experiments across SWOKS and TFCL.}\vspace{8pt}
    \label{tab.ppohyperparameters}
    \end{center}
    \end{table}

    \begin{table}
    \centering
    \begin{center}
    \small
        \begin{tabular}{|p{4cm}|c|c|c|c|}
        \hline
        Layer & Channel & Kernel & Stride & Padding \\
        \hline
        Input & 3 & - & - & -\\
        Conv1 & 16 & (2x2) & 1 & 0 \\
        ReLU & - & - & - & - \\
        Max Pool 2-d & 16 & (2x2) & 2 & 0 \\
        Conv2 & 32 & (2x2) & 1 & 0 \\
        ReLU & - & - & - & - \\
        Conv3 & 64 & (2x2) & 1 & 0 \\
        ReLU & - & - & - & - \\
        Flatten & - & - & - & - \\
        Linear & 200 & - & - & - \\
        ReLU & - & - & - & - \\
        \hline
        \end{tabular}
        \caption{Network model in PPO for Minigrid experiments.}
        \label{tab.model}
    \end{center}
    \end{table}

SWOKS, MBCD and 3RL are employed with an SAC network, with parameters as follows. MBCD is used without simulated replay to work under the same assumptions on input data as SWOKS and 3RL. The hyperparameters used for the SAC network is given in Table~\ref{tab.sachyperparameters}.
   \begin{table}[h]
   \begin{center}
   \begin{tabular}{|p{5cm}|p{2.5cm}|p{2.5cm}|p{2.5cm}|}
   \hline
   Hyper-Parameter & Half-Cheetah \\
   \hline
   Learning rate & 0.00015 \\
   CL preservation & masks \\
   Number of workers & 4 \\
   Optimizer function & RMSprop \\
   Discount rate & 0.99 \\
   Entropy weight & 0.00015 \\
   Rollout length & 128 \\
   Optimization epochs & 8 \\
   Number of mini-batches & 64 \\
   PPO ratio clip & 0.1 \\
   Iteration log interval & 1 \\
   Gradient clip & 5 \\
   Max steps (per task) & 12800 \\
   Evaluation episodes & 25 \\
   Require task label & True \\
   Backbone network seed & 9157 \\
   \hline
   \end{tabular}
   \caption{SAC hyper-parameters used for Half-Cheetah experiments across SWOKS, MBCD, and 3RL.}\vspace{8pt}
   \label{tab.sachyperparameters}
   \end{center}
   \end{table}

 \subsection{CT-graph environment hyper-parameters}
    \begin{table}[H]
    \begin{center}
    \small
    \begin{tabular}{|p{4.9cm}|p{3.8cm}|}
    \hline
    Hyper-Parameter & Depth 2 CT-graph (4 tasks) \\
    \hline
    General seed & 1\\
    Tree depth & 2\\
    Branching factor & 2 \\
    Wait probability & 0.0 \\
    High reward value & 1.0 \\
    Fail reward value & -0.1 \\
    Stochastic sampling & false  \\
    Reward standard deviation & 0  \\
    Min static reward episodes & 0 \\
    Max static reward episodes & 0 \\
    Reward distribution & needle in haystack \\
    MDP decision states & true  \\
    MDP wait states & true  \\
    Wait states & $[2,8]$ \\
    Decision states & $[9,11]$  \\
    Graph ends & $[12,15]$ \\
    Image dataset seed & 1  \\
    1D format & false  \\
    Number of images & 16  \\
    Noise on images on read & 0 \\
    Small rotation on read & 1 \\
    \hline
    \end{tabular}
    \caption{CT-graph environment parameters. We use CT-graph to compare SWOKS to TFCL.}\vspace{8pt}
    \label{tab.ctparams}
    \end{center}
    \end{table}

  \subsection{Minigrid environment hyper-parameters}
\begin{figure}
 \centering
   \includegraphics[width=0.2\textwidth]{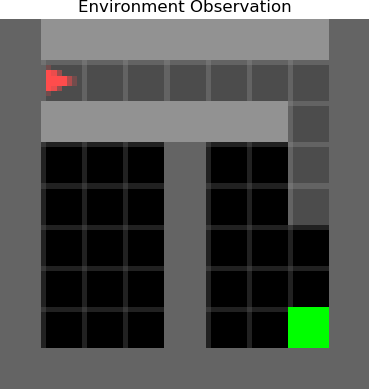}
   \includegraphics[width=0.2\textwidth]{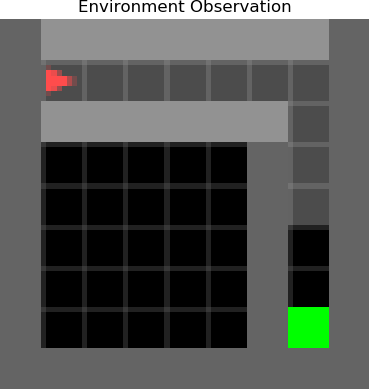}
   \includegraphics[width=0.2\textwidth]{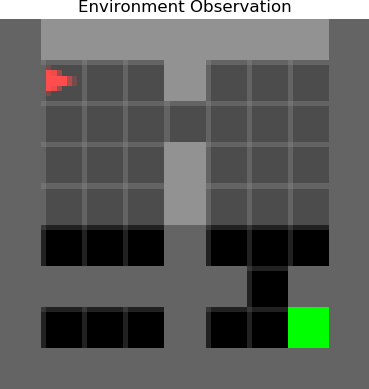}
   \includegraphics[width=0.25\textwidth]{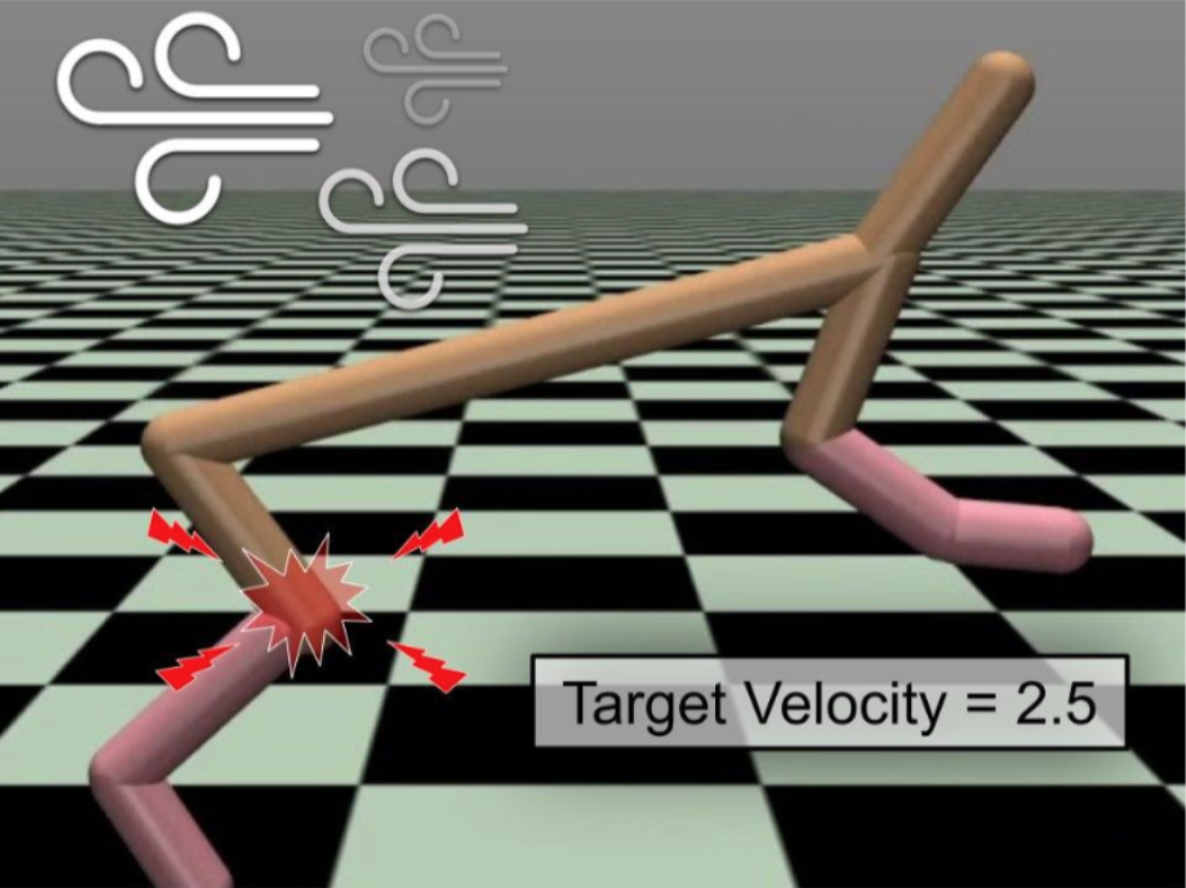}
   \caption{Left: Illustrations of the Minigrid SimpleCrossingS9N2-v0 task variations used in experiments. Right: Dramatized illustration of task changes in the Half-Cheetah experiments (Half-Cheetah figure from \citet{alegre2021minimum}).}
   \label{fig:mngr}
 \end{figure}

  Table~\ref{tab.mgparams} contains the environment hyper-parameters used for experiments in the Minigrid environment. Three tasks are used within the environment, with different walls placed in different locations for different tasks (Figure~\ref{fig:mngr}). As the environment is partially observable, some states will look the same between tasks, making different tasks appear very similar. Seeds used are the same as in \citet{kessler2022same}.
    \begin{table}
    \begin{center}
    \begin{tabular}{|p{4cm}|p{6cm}|}
    \hline
    Hyper-Parameter & Minigrid SC (3 tasks) \\
    \hline
    Tasks & MiniGrid-SimpleCrossingS9N2-v0 \\
    Seeds & 129, 112, 111 \\
    \hline
    \end{tabular}
    \caption{Minigrid environment parameters. We use Minigrid to compare SWOKS to TFCL.}\vspace{8pt}
    \label{tab.mgparams}
    \end{center}
    \end{table}

\subsection{Half-Cheetah environment hyperparameters}
Tasks experienced by Half-Cheetah are: Normal, Joint-Malfunction, Velocity, and Wind. These tasks are dramatically visualized in Figure~\ref{fig:mngr}, and summarized in Table~\ref{tab.halfcheetahparams}.

   \begin{table}
   \begin{center}
   \begin{tabular}{|p{4cm}|p{7cm}|}
   \hline
   Hyper-Parameter & Half-Cheetah (4-tasks) \\
   \hline
   Tasks & normal, joint-malfunction, wind, velocity \\
   Change freq. & 40000 \\
   Default target vel. & 1.5 \\
   Velocity target vel. & 2.0 \\
   Joint malfunction mask & [-1, -1, 0, 0, 0, 0] \\
   \hline
   \end{tabular}
   \caption{Half-cheetah environment parameters. We use Half-Cheetah to compare SWOKS to MBCD and 3RL.}\vspace{8pt}
   \label{tab.halfcheetahparams}
   \end{center}
   \end{table}

\subsection{Implementation Libraries}

Implementation libraries are different between experiments due to the differing underlying networks implemented. Half-Cheetah experiments utilize TensorFlow, while the CT-graph and Minigrid experiments use pytorch. Package libraries for the different experiments conducted are in Table~\ref{tab.packages}.

    \begin{table}
    \begin{center}
    \begin{tabular}{|p{4.5cm}|p{4cm}|p{4cm}|}
    \hline
    Package & CT-graph \& Minigrid & Half-Cheetah \\
    \hline
    python & 3.7.0 & 3.7.16 \\
    pytorch & 1.12.1 & n/a \\
    gym & 0.26.2 & 0.19.0 \\
    gymnasium & 0.28.1 & 0.28.1 \\
    minigrid & 2.3.0 & n/a\\
    gym-ctgraph & 1.0 & n/a\\
    cuda toolkit & 11.6.0 & 10.0.130\\
    cudnn & n/a & 7.6.5\\ 
    cython & 0.29.36 & 0.29.36\\
    pot & 0.9.1 & 0.9.1\\
    tensorflow & n/a & 1.15.0\\
    mujoco-py & n/a & 2.1.2.14 \\
    
    \hline
    \end{tabular}
    \vspace{8pt}
    \caption{Packages used in the SWOKS implementation with Minigrid and CT-graph. The system is based on a PyTorch RL implementation using CUDA and Gym as the environment interface. Calculations of the sliced Wasserstein distances are performed using the Python Optimal Transport library \citep{flamary2021pot}. We use Gymnasium for CT-graph and Minigrid and Gym for Half-Cheetah.}
    \label{tab.packages}
    \end{center}
    \end{table}

\end{document}